# Facial Expressions recognition Based on Principal Component Analysis (PCA)

Abdelmajid Hassan Mansour[#1], Gafar Zen Alabdeen Salh[*2], Ali Shaif Alhalemi[#3]

[#]1 Assistant Professor, Faculty of Computers and Information Technology, King Abdulaziz University, Saudi Arabia
[*]2 Assistant Professor, Faculty of Computers and Information Technology, King Abdulaziz University, Saudi Arabia
[#]3 Lecturer, Faculty of Computer Science and Information Technology, Alneelain University, Sudan

*Abstract*— The facial expression recognition is an ocular task that can be performed without human discomfort, is really a speedily growing on the computer research field. There are many applications and programs uses facial expression to evaluate human character, judgment, feelings, and viewpoint. The process of rrecognizing facial expression is a hard task due to the several circumstances such as facial occlusions, face shape, illumination, face colors, and etc. This paper present a PCA methodology to distinguish expressions of faces under different circumstances and identifying it. Relies on Eigen faces technique using standard Data base images. So as to overcome the problem of difficulty to computers to identify the features and expressions of persons.

## I. INTRODUCTION

Facial expression recognition is easy task for humans not for computer. So it can be played a vital role in human-computer interaction. Human beings naturally use facial expression to communicate their emotions and to interact socially. In human face the features of the face such as lip corner, eye corners and nose tip are critical points [2].

Expression is an important mode of non-verbal conversation among people. Recently, the facial expression recognition technology attracts more and more attention with people's growing interesting in expression information. Facial expression provides essential information about the mental, emotive and in many cases even physical states of the conversation. Face expression recognition possesses practically significant importance, it offers vast application prospects, such as user-friendly interface between people and machine, humanistic design of products, and an automatic robot for example. Face perception is an important component of human knowledge. Faces contain much information about ones id and also about mood and state of mind. Facial expression interactions usually relevant in social life, teacher-student interaction, credibility in numerous contexts, medicine etc. however people can easily recognize facial expression easily, but it is quite hard for a machine to do this [1].

Face recognition is a difficult problem because of the generally similar shape of faces combined with the numerous variations between images of the same face. The image of a face changes with facial expression, age, viewpoint, illumination conditions and noise etc. The task of a face recognition system is to recognize a face in a manner that is as independent as possible of these image variations.

Automatic recognition of faces is considered as one of the fundamental problems in computer vision and pattern analysis, and many scientists from different areas have addressed it [3].

In this paper we present a PCA methodology to distinguish expressions of the faces under different circumstances and identifying it. So as to overcome the problem of difficulty to computers to identify the features and expressions of persons.

## II. PRINCIPAL COMPONENT ANALYSIS (PCA)

PCA is a technique of identifying patterns in data, and expressing the data in such a way so as to highlight their differences and similarities. (PCA) involves some sort of numerical procedure that changes several (possibly) correlated variables into a smaller number of uncorrelated variables called principal components [1]. PCA is the most statistical method that used, while retaining the informative variables, thus making it easier to operate the data. PCA manages the entire data for the principal components analysis without taking into consideration the fundamental class structure [2].

The PCA is a technique used to reduce the dimensionality which can be used to solve compression and recognition problems, is also known as Hotelling, or eigenspace Projection or Karhunen and Leove (KL) transformation [5]. In PCA the original data image is transformed into a subspace set of Principal Components (PCs) such that the first orthogonal dimension of this subspace captures the greatest amount of variance among the images. The last dimension of this subspace captures the least amount of variance among the image, based on the statistical characteristics of the targets [6].

PCA has the potential to perform feature extraction, that able to capture the most variable data components of samples, and select a number of important individuals from all the feature components. In the field of face recognition, image denoising, data compression, data mining, and machine learning PCA has been successfully used. Implementation of the PCA method in face recognition is called eigenfaces technique [7].

## III. RELATED WORK

Facial Expressions recognition is one of the most powerful, natural, and immediate means for human beings to communicate their emotions and intentions. There are several studies about the recognition of Facial Expressions. Saket S

Kulkarni, Narender P Reddy and SI Hariharan[8], was developed an intelligent system for facial image based expression classification using committee neural networks, The system correctly identified the correct facial expression in 255 of the 282 images (90.43% of the cases), from 62 subjects not used in training or in initial testing. Reviews various techniques of facial expression recognition systems using MATLAB (neural network) toolbox. Was introduced by Pushpaja V. Saudagare, D.S. Chaudhari [9]. Nancy Smith [10], demonstrates that PCA eigenvectors are able to represent the patterns unique to basic emotions across different faces. Rehmat Khan, Rohit Raja [11], was introduced description schemes for selecting the image and then processing the image to recognize the expressions.

### IV. PROPOSED SCHEME

The proposed work uses PCA technology that characterized by simplicity and effectiveness in the use of time and memory space. Depending on the concepts of artificial intelligence and image processing. So as to enable the computer to identify the features and expressions of the faces of the people and to distinguish them according to the biometrics for making the right decision. Where certain image can be classified into one of the seven basic categories of facial expressions (Happiness, Sadness, Fear, Surprise, Anger, Disgust and Neutral). These categories was considered as the basic factors of ensuring the ability of computer's to recognize facial expressions in different circumstances.

The proposed system uses four databases, three of them standard templates image databases (Utrecht database collected at the European Conference of visual perception in Utrecht, Pain expressions databases that were created at the University of North Columbia British contains a new expression called Pain expression, used in the evaluation of pain through facial expressions, Indian database created at Indian university, and in addition to the database created by the researcher by taking pictures under various circumstances according to the previous categories.

The system loads the image that have been recently captured and then analyze facial expressions using PCA technique and compare it with images already stored in advance on the system database in order to identify the expression that the person holds it through different visual inputs. The proposed system move through four stages for identifying the facial expressions, as shown in Fig. 1.

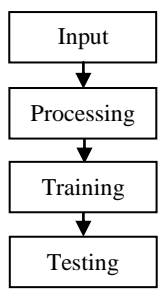

Fig. 1. Identifying the facial expressions steps

### V. TRAINING

The proposed system will be trained on the four databases, as follow:

*1) Utrecht database*: Consists of 131 images, divided into 49 males and 20 females, the images contain smiling and natural expressions, with dimensions of 900 * 1200.

*2) Indian database*: Consists of 50 male images with different expressions as (13 image bearing the expression happy, 11 bearing disgust, 10 bearing angry, 9 bearing sad, and 7 image bearing the expression neutral, as shown in Fig. 2.

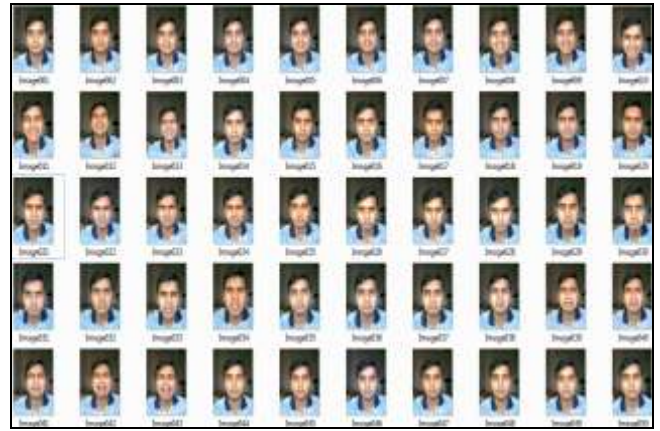

Fig. 2. Indian Database

*3) Researcher Database:* Consists of 45 images from three different persons with different expressions as (7 neutral, 6 happy, and 7 disgust, 12 surprise, 8 anger, and 5 images bearing the expression sad, these images were taken by Iphone4 Mobile camera accurately 5 Megapixels in a medium lighting, as shown in Fig. 3.

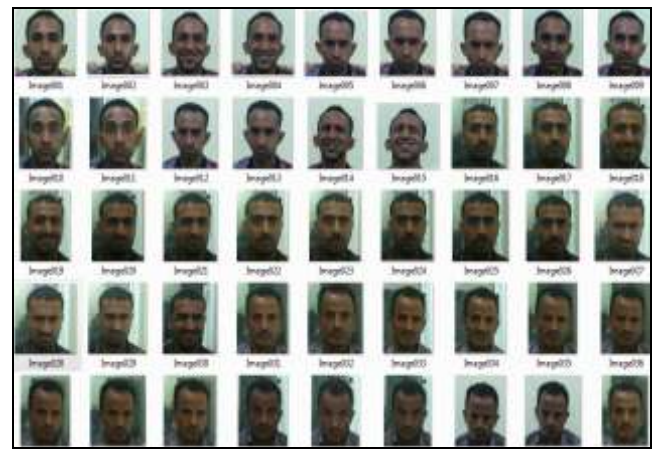

Fig. 3. Researcher Database

*4) Pain expressions database*: Consists of 599 image categorized into (13 females have 339 image, and 10 males have 260 image), there are about 26 image for every person whether male or female, distributed into 9 different expressions as ( 2 image bearing the expression change of direction (rotation) by 45 degree and another by 90 degree, 2 have angry, 2 have disgust, 2 have fear, 2 have happy, 2 have neutral, 2 have surprise, 2 have sad, and 10 images have the pain expression. Where the image resolution is 576 * 720. The process of data training for the four databases shown in Fig. 4.

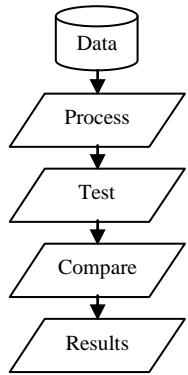

Fig. 4. Data Training Process for Facial Expression using PCA

## VI. TESTING PHASE

This stage perform the extraction of the face features from image, the same as the method used in the training phase. The result is an index of emotion for the image of the face, which are subject to the testing process, as shown in Fig. 5.

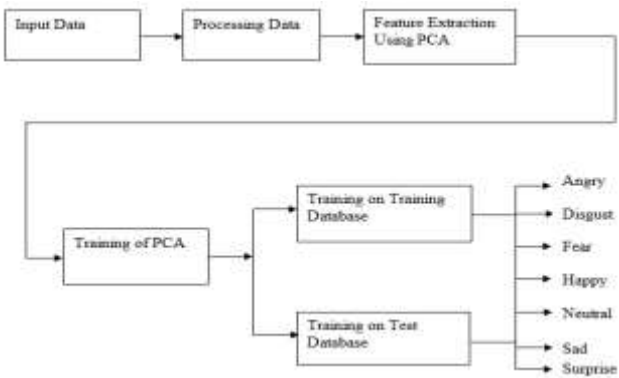

Fig. 5. Testing Process of identifying the emotion in general

### A. Testing Indian Database

Testing Indian Database for Males: The result of testing Indian Database process the system classified 29 image out of 31 image, as shown in Table I.

Table I. Results of testing Indian Database for "Males"

| total image | feeling | tested image | true classify | false classify | true rate % | false rate % |
|---|---|---|---|---|---|---|
| 13 | happy | 7 | 6 | 1 | 85.7 | 14.3 |
| 11 | disgust | 8 | 7 | 1 | 87.5 | 12.5 |
| 10 | anger | 6 | 6 | 0 | 100 | 0 |
| 9 | Sad | 7 | 7 | 0 | 100 | 0 |
| 7 | neutral | 3 | 3 | 0 | 100 | 0 |
| Total | | 31 | 29 | 2 | %93.5 | %6.5 |

The result of testing Indian Database for Males is represented in chart diagram as shown in Fig. 6.

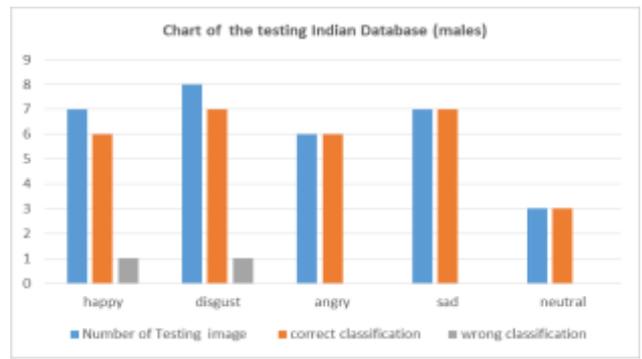

Fig. 6. Chart of testing Indian Database for "Male"

### B. Testing pain expression Database

*1) Testing pain expression Database for "Male"*: The result of testing Pain expression Database process for male's image, the system classified 257 image out of 260 image, as shown in Table II.

Table II. Results of testing Pain expression Database for "Males"

| total image | feeling | tested image | true classify | false classify | true rate % | false rate % |
|---|---|---|---|---|---|---|
| 20 | rotation | 20 | 20 | 0 | 100 | 0 |
| 20 | anger | 20 | 19 | 1 | 95 | 5 |
| 20 | disgust | 20 | 20 | 0 | 100 | 0 |
| 20 | Fear | 20 | 20 | 0 | 100 | 0 |
| 20 | happy | 20 | 20 | 0 | 100 | 0 |
| 20 | neutral | 20 | 20 | 0 | 100 | 0 |
| 100 | pain | 100 | 98 | 2 | 98 | 2 |
| 20 | surprise | 20 | 20 | 0 | 100 | 0 |
| 20 | sad | 20 | 20 | 0 | 100 | 0 |
| Total | | 260 | 257 | 3 | 99.22% | 0.78% |

The result of testing pain expression Database for male's image, is represented in chart diagram as shown in Fig. 7.

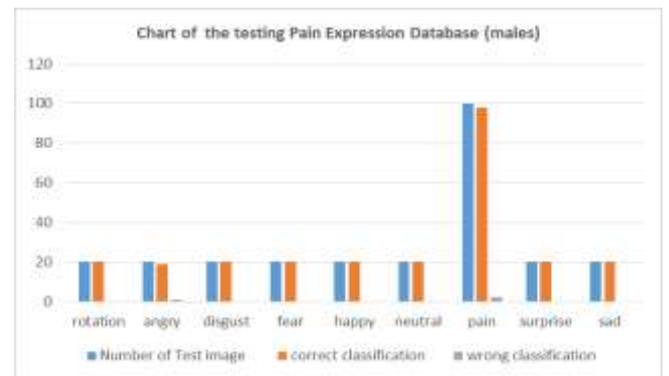

Fig. 7. Chart of testing Pain expression Database for "Male"

*1) Testing pain expression Database for "Female"*: After the testing process the system classified 336 image out of 339 image from 13 (each one has 26 image, except one has 27 image), as shown in Table III.

Table III. Results of testing Pain expression Database for "Females"

| total image | feeling | tested image | true classify | false classify | true rate % | false rate % |
|---|---|---|---|---|---|---|
| 27 | rotation | 27 | 27 | 0 | 100 | 0 |
| 26 | anger | 26 | 26 | 0 | 100 | 0 |
| 26 | disgust | 26 | 26 | 0 | 100 | 0 |
| 26 | Fear | 26 | 26 | 0 | 100 | 0 |
| 26 | happy | 26 | 26 | 0 | 100 | 0 |
| 26 | neutral | 26 | 26 | 0 | 100 | 0 |
| 130 | pain | 130 | 127 | 3 | 97.6 | 2.4 |
| 26 | surprise | 26 | 26 | 0 | 100 | 0 |
| 26 | sad | 26 | 26 | 0 | 100 | 0 |
| | Total | 339 | 336 | 3 | 99.7% | 0.26% |

The result of testing pain expression Database for Females image, is represented in chart diagram as shown in Fig. 8.

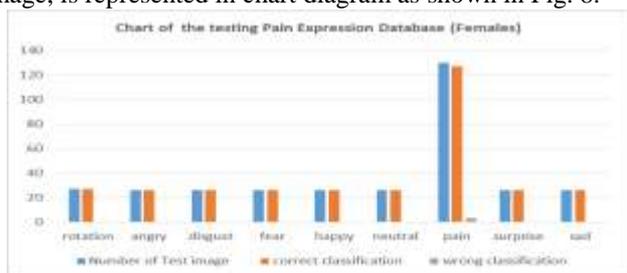

Fig. 8. Chart of testing Pain expression Database for "Female"

*2) Testing pain expression Database for "Female and Males"*: After the testing Pain expression Database the system classified 593 image out of 599 image, as shown in Table IV.

Table IV. Results of testing Pain expression Database for "Females and Males"

| total image | feeling | tested image | true classify | false classify | true rate % | false rate % |
|---|---|---|---|---|---|---|
| 27 | rotation | 47 | 47 | 0 | 100 | 0 |
| 26 | anger | 46 | 45 | 1 | 97.8 | 2.2 |
| 26 | disgust | 46 | 46 | 0 | 100 | 0 |
| 26 | Fear | 46 | 46 | 0 | 100 | 0 |
| 26 | happy | 46 | 46 | 0 | 100 | 0 |
| 26 | neutral | 46 | 46 | 0 | 100 | 0 |
| 130 | pain | 230 | 225 | 5 | 97.8 | 2.2 |
| 26 | surprise | 46 | 46 | 0 | 100 | 0 |
| 26 | sad | 46 | 46 | 0 | 100 | 0 |
| | Total | 599 | 593 | 6 | 98.9% | 1.1% |

The result of testing pain expression Database for Females and males, is represented in chart diagram as shown in Fig. 9.

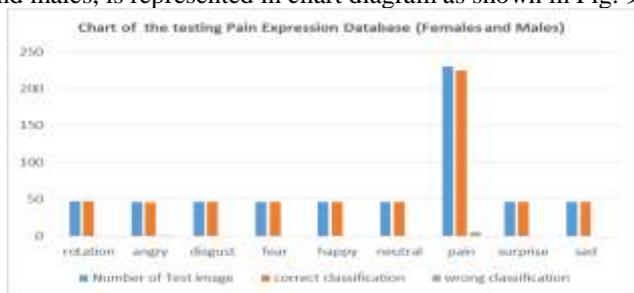

Fig. 9. Chart of testing Pain expression Database for "Female and Male"

*C. Testing Utrecht Database*

*1) Testing Utrecht Database for Males*: The results of Testing Utrecht Database process, the system classified all 47 image, as shown in Table V.

Table V. Results of testing Utrecht Database for "Males"

| Total image | feeling | tested image | true classify | false classify | true rate % | false rate % |
|---|---|---|---|---|---|---|
| 52 | neutral | 27 | 27 | 0 | 100 | 0 |
| 40 | happy | 20 | 20 | 0 | 100 | 0 |
| | Total | 47 | 47 | 0 | 100% | 0% |

The result of testing Utrecht Database for Males image, is represented in chart diagram as shown in Fig. 10.

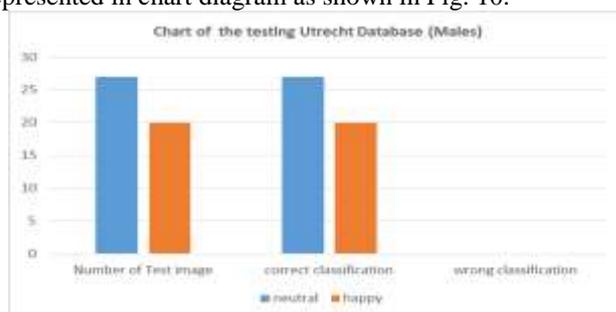

Fig. 10. Chart of testing Utrecht Database for "Male"

*2) Testing Utrecht Database for Females*: The results of Testing Utrecht Database process the system classified all 20 image of Females, as shown in Table VI.

Table VI. Results of testing Utrecht Database for "Females"

| Total image | feeling | tested image | true classify | false classify | true rate % | false rate % |
|---|---|---|---|---|---|---|
| 17 | neutral | 7 | 7 | 0 | 100 | 0 |
| 22 | happy | 13 | 13 | 0 | 100 | 0 |
| | Total | 20 | 20 | 0 | 100% | 0% |

The result of testing Utrecht Database for Females image, is represented in chart diagram as shown in Fig. 11.

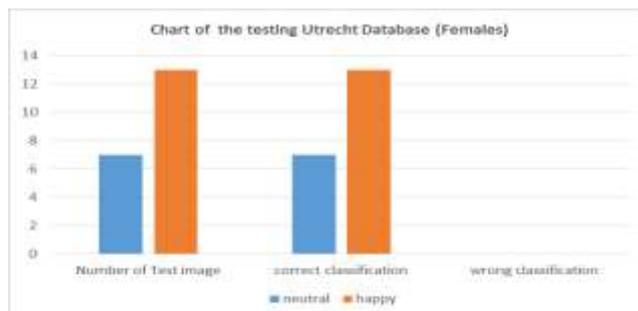

Fig. 11. Chart of testing Utrecht Database for "Female"

*3) Testing Utrecht Database for Females and Males*: The results of Testing Utrecht Database process the system classified all 67 image of Females and males, as shown in Table VII.

Table VII. Results of testing Utrecht Database for "Females and Male"

| Total image | feeling | tested image | true classify | false classify | true rate % | false rate % |
|---|---|---|---|---|---|---|
| Male | neutral | 27 | 27 | 0 | 100 | 0 |
| Female | neutral | 7 | 7 | 0 | 100 | 0 |
| Male | happy | 20 | 20 | 0 | 100 | 0 |
| Female | happy | 13 | 13 | 0 | 100 | 0 |
| Total | | 67 | 67 | 0 | 100% | 0% |

The results of testing Utrecht Database for both Females and Males image, is represented in chart diagram as shown in Fig. 12.

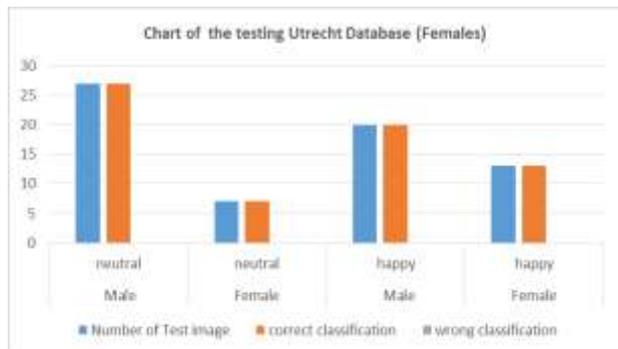

Fig. 12. Chart of testing Utrecht Database for "Females and Males"

*D. Testing Researcher Database*

The result of testing Researcher Database process, the system classified 44 image out of 45 image for Males, as shown in Table VIII.

Table VIII. Results of testing Researcher Database for "Males"

| Total image | feeling | tested image | true classify | false classify | true rate % | false rate % |
|---|---|---|---|---|---|---|
| 7 | neutral | 7 | 7 | 0 | 100 | 0 |
| 6 | happy | 6 | 6 | 0 | 100 | 0 |
| 7 | disgust | 7 | 7 | 0 | 100 | 0 |
| 12 | surprise | 12 | 12 | 0 | 100 | 0 |
| 8 | anger | 8 | 7 | 1 | 87.5 | 12.5 |
| 5 | sad | 5 | 5 | 0 | 100 | 0 |
| Total | | 45 | 44 | 0 | 97.8% | 2.2% |

The results of testing Researcher Database for Males image, is represented in chart diagram as shown in Fig. 13.

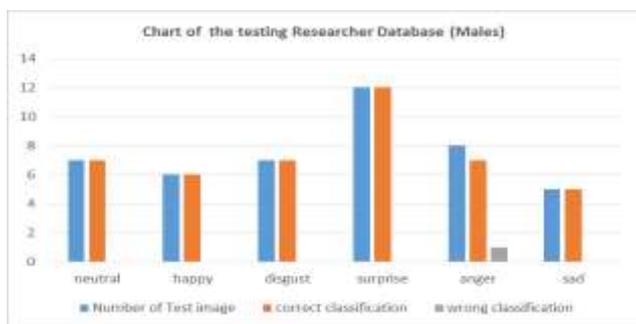

Fig. 13. Chart of testing Researcher Database for "Males"

*E. Comparing all Databases*

The final results of comparing the Test of the 4 Databases process the system classified 409 image of 451 image for Females and Males, as shown in the following table:

Table IX. Results of testing the 4 Databases for "Female and Males"

| Data Base | Total Image | tested image | true classify | false classify | true rate % | false rate % |
|---|---|---|---|---|---|---|
| Indian | 50 | 31 | 29 | 2 | 93.5 | 6.5 |
| Pain expression | 599 | 599 | 593 | 6 | 98.9 | 1.1 |
| Utrecht | 131 | 67 | 67 | 0 | 100 | 0 |
| Researcher | 45 | 45 | 44 | 1 | 97.7 | 2.3 |
| Total | 825 | 742 | 733 | 9 | 98.78% | 1.22% |

The results of testing the 4 Databases for both Females and Males, is represented in chart diagram, as shown in Fig. 13.

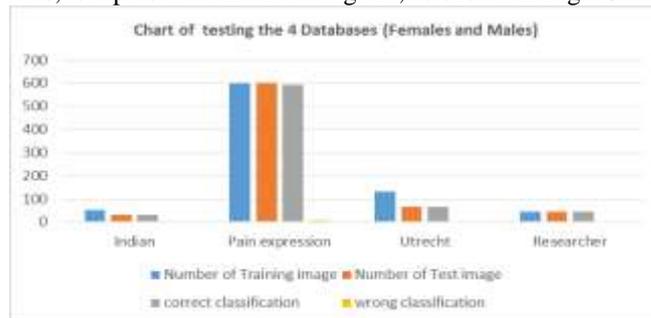

Fig. 14. Chart of testing Researcher Database for "Males"

## VII. CONCLUSION

The system was trained on 825 image (male and female) vary in degree of lighting, direction of the face, and facial expressions. While the system was tested on 742 image, and identified 733 different expressions from them with rate 98.78%. this means the results of experiments that applied to the proposed system was good and ensure that the methods used to recognize and extract the expressions of the faces is a good, and their accuracy of classification is high according to the techniques currently available.

There are some factors (such as the type of camera used, distance from camera, and lighting) may affect the process of recognition. Increase the training data of the system lead to increase training time. The Utrecht database gave the best results and very high rating in the classification. The PCA algorithm proved that with high efficiency in the process of facial expressions recognition in less time.

## AUTHOR PROFILE

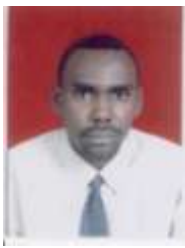
**Dr. Abdelmajid Hassan Mansour Emam**, Assistant Professor, Department of Computers and Information Technology, King Abdulaziz University, Faculty of Computers and Information Technology, khulais, Jeddah, Saudi Arabia..
**Permanent Address**: Department of Information Technology, Faculty of computer Science and Information Technology, Alneelain University, Sudan,

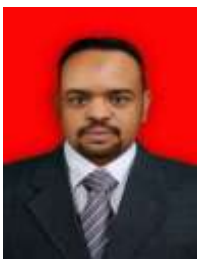
**Dr. Gafar Zen Alabdeen Salh Hassan**, Assistant Professor, Department of Computers and Information Technology, King Abdulaziz University, Faculty of Computers and Information Technology, khulais, Jeddah, Saudi Arabia..
**Permanent Address**: Department of Information Technology, Faculty of computer Science and Information Technology, Alneelain University, Sudan,

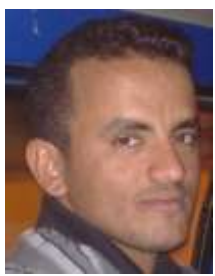
Ali Shaif Alhalemi, Lecturer, Department of Information Technology, Faculty of computer Science and Information Technology, Alneelain University, Sudan,